%% file: submission.tex
\documentclass[runningheads]{llncs}
\usepackage[T1]{fontenc}
\usepackage{graphicx}

\usepackage[utf8]{inputenc} 
\usepackage{amsfonts,amssymb,amsmath,bm}
\usepackage{booktabs}
\usepackage[sectionbib,sort&compress]{natbib}
\usepackage{multirow}
\usepackage{hyperref}

\usepackage{pgfplots}
\DeclareUnicodeCharacter{2212}{−}
\usepgfplotslibrary{groupplots,dateplot}
\usetikzlibrary{patterns,shapes.arrows,shapes.geometric}
\pgfplotsset{compat=newest}

\titlerunning{Measuring Generalization Performance Across Multiple Objectives}
\title{Mind the Gap: Measuring Generalization Performance Across Multiple Objectives 
}

\author{%
  Matthias Feurer\inst{1} \and  Katharina Eggensperger\inst{1} \and Edward Bergman\inst{1} \and \\ Florian Pfisterer\inst{2,3} \and Bernd Bischl\inst{2,3} \and Frank Hutter\inst{1,4}
}

\institute{Albert-Ludwigs-Universität Freiburg \and Ludwig-Maximilians-Universität München \and Munich Center for Machine Learning \and Bosch Center for Artificial Intelligence \\ \email{\{feurerm,eggenspk,bergmane,fh\}@cs.uni-freiburg.de} \\  \email{\{florian.pfisterer,bernd.bischl\}@stat.uni-muenchen.de}
}
\authorrunning{Feurer, Eggensperger, Bergman, Pfisterer, Bischl and Hutter}

\begin{document}

\include{macros}

\maketitle

\begin{abstract}
Modern machine learning models are often constructed taking into account multiple objectives, e.g.,\ minimizing inference time while also maximizing accuracy. Multi-objective hyperparameter optimization (\mhpo{}) algorithms return such candidate models, and the approximation of the Pareto front is used to assess their performance. 
In practice, we also want to measure generalization when moving from the validation to the test set. However, some of the models might no longer be Pareto-optimal which makes it unclear how to quantify the performance of the \mhpo{} method when evaluated on the test set.
To resolve this, we provide a novel evaluation protocol that allows measuring the generalization performance of \mhpo{} methods and studying its capabilities for comparing two optimization experiments.
\end{abstract}

\section{Introduction}

Multi-objective hyperparameter optimization (\mhpo{};~\citealp{feurer-automlbook19a,moraleshernandez-arxiv21a,karl-arxiv22a}) and multi-objective neural architecture search (MNAS;~\citealp{elsken-jmlr19a,benmeziane-arxiv21a}) are becoming increasingly important and enable moving beyond the purely performance-driven selection of machine learning (ML) models.
Important additional objectives are, for example, model size, inference time and the number of operations~\citep{elsken-iclr19a}, interpretability~\citep{molnar-ecml19a}, feature sparseness~\citep{binder-gecco20a}, or fairness~\citep{chakraborty-ase19a,cruz-icdm21a,schmucker-arxiv21a}.
To evaluate and compare multi-objective methods, papers often report the hypervolume indicator of the Pareto front approximation as a measure of optimization performance.

However, as we show in this paper, an ML model that is located on the approximated Pareto front on the validation set can become a dominated model on the test set and vice versa. This phenomenon makes it impossible to compute the hypervolume indicator using the canonical \emph{train-validation-test} evaluation protocol~\citep{raschka-arxiv18a}.
To remedy this, we propose a novel evaluation protocol that takes such models into account in order to lay a solid foundation for multi-objective hyperparameter optimization.
In addition, we also conduct an initial study in which we use this evaluation protocol to compare the hyperparameter optimization of two machine learning algorithms.

This paper is structured as follows. First, in Section~\ref{sec:mooeval}, we give background on multi-objective optimization. In Section~\ref{sec:measurefairnesstest}, we then discuss the problem of evaluating generalization performance on a test set and the problems of a naive solution. We go on and describe our new protocol in Section~\ref{sec:newprotocol} and exemplify it in Section~\ref{sec:experiment}. Then, 
we describe how multi-objective generalization was (not) measured in related work in Section~\ref{sec:prior} before concluding the paper in Section~\ref{sec:conclusion}.

We provide Python code to reproduce our experiments at \newline{} \href{https://github.com/automl/IDA23-MindTheGap}{https://github.com/automl/IDA23-MindTheGap}.

\section{Background}\label{sec:mooeval}

In the remainder of this paper, we follow the notation from \citet{karl-arxiv22a} and aim to minimize the multi-objective function $\vec{c} : \configspacelambda \rightarrow \mathbb{R}^\metricdim$ defined as $\min_{\conf \in \configspacelambda} \vec{\cost}(\conf) = \min_{\conf \in \configspacelambda} (\cost_1(\conf), \dots, \cost_\metricdim(\conf))$, where each $\cost_i : \configspacelambda \rightarrow \mathbb{R}$ denotes the cost of hyperparameter configuration (HPC) $\conf \in \configspacelambda$ according to one cost metric $i \in (1, \dots, \metricdim)$. 
Since, typically, there is no total order on the space of objectives $\mathbb{R}^M$, and hence there usually is no single best objective value, we now consider \emph{Pareto-dominance} and \emph{Pareto-optimality} instead.
Given a function $\vec{c}: \configspacelambda \rightarrow \mathbb{R}^M$, we define a binary relation $`\prec`$ on $\mathbb{R}^M \times\mathbb{R}^M$. Given two cost vectors $\val^{(1)}, \val^{(2)} \in \mathbb{R}^M$, defined as $\val^{(1)} = \vec{c}(\conf^{(1)})$ and $\val^{(2)} = \vec{c}(\conf^{(2)})$, we say $\val^{(1)}$ \emph{dominates} $\val^{(2)}$, written as $\val^{(1)} \prec \val^{(2)}$, if and only if
\begin{align*}
 \forall k \in \left\{1, ..., M\right\}: \val_k^{(1)} \leq \val_k^{(2)} \; \land \;
 \exists l \in \left\{1, ..., M\right\}: \val_l^{(1)}  <  \val_l^{(2)}.
\end{align*}
We similarly define a dominance relationship for configurations $\conf$: A configuration $\conf^{(1)}$ dominates another configuration $\conf^{(2)}$, so $\conf^{(1)} \prec \conf^{(2)}$. if and only if $\vec{c}(\conf^{(1)}) \prec \vec{c}(\conf^{(2)})$.
The \emph{non-dominated} set of solutions, the Pareto front $\mathcal{P}$, is then given by 
$\mathcal{P} = \{\val \in \vec{c}(\configspacelambda) \,|\, \nexists \, \val' \in \vec{c}(\configspacelambda) \text{ s.t. } \val'\prec \val \}$ and conversely, the Pareto set as the pre-image of $\mathcal{P}$: $\mathcal{P}_{\configspacelambda} = \vec{c}^{-1}(\mathcal{P}) = \{\conf \in \configspacelambda \,|\, \nexists \, \conf' \in \configspacelambda \text{ s.t. } \conf'\prec \conf \}$.

An \mhpo{} algorithm then aims to return the best approximation of the Pareto front, trading off all given objectives. To obtain candidate HPCs $\conf^{(i)}$, an \mhpo{} algorithm iteratively generates and evaluates HPCs $\{\conf^{(1)}, \conf^{(2)}, \dots, \conf^{(T)}\}$. In the next step, the \mhpo{} algorithm\footnote{In principle, this is agnostic to the capability of the HPO algorithm to consider multiple objectives. Any HPO algorithm (including random search) would suffice since one can compute the Pareto-optimal set post-hoc.} compares the performance of all evaluated solutions to obtain the subset of HPCs $\tilde{P}_{\configspacelambda} \subseteq \{\conf^{(1)}, \conf^{(2)}, \dots, \conf^{(T)}\}$ approximating the Pareto set and thereby also the Pareto front.
\footnote{The true Pareto front is only \textit{approximated} because there is usually no guarantee that an \mhpo{} algorithm finds the optimal solution. Furthermore, there is no guarantee that an algorithm can find all solutions on the true Pareto front.}

The literature provides several quality metrics for Pareto-optimal sets focusing on different aspects~\citep{zitzler-ec00a,zitzler-tec03a,emmerich-naturecomp18a}: (1) Approximation quality of the Pareto front, (2) a good (often uniform) distribution of solutions, and (3) diversity \wrt{} to the values for each metric. Here, we consider the commonly used hypervolume indicator~\citep{karl-arxiv22a}, i.e., the volume of the objective space covered by the dominating solutions \wrt{} a reference point. The hypervolume indicator mostly considers (1), and it can be used to capture the performance of an \mhpo{} experiment in a single value. 

While we discuss our work in the context of \mhpo{}, the background, problem, and proposed solution also apply to other multi-objective optimization problems which involve separate validation and test sets, such as neural architecture search (NAS;~\citealp{elsken-jmlr19a}), Automated Machine Learning (\automl{};~\citealp{hutter-book19a}), or ensemble learning.

\section{Evaluating Generalization}\label{sec:measurefairnesstest}

Having discussed how to evaluate an \mhpo{} method in general, we now turn to the problem that in ML, the predictive performance is usually measured on unseen test data. To highlight the challenges, we summarize the standard evaluation protocol, describe a previously unknown failure mode, hypothesize a naive solution and point out two potential issues of such a naive solution.

In \mhpo{}, we typically tune the hyperparameters of an ML model on a supervised ML task, e.g., classification, with dataset $\dataset = [(\inp, \out)^{(1)}, \dots, (\inp, \out)^{(d)}]$.
We consider minimizing data-based costs $c_i$ that estimate an empirical risk \wrt{} to the entire data distribution, e.g., the empirical risk, fairness metrics, or explainability scores (in contrast to model-based costs, such as inference time or model size). Because we only have access to a finite sample from the entire data distribution, we estimate this risk using the canonical \emph{train-validation-test} protocol~\citep{raschka-arxiv18a}, which trains models on the \emph{train} portion of the data, and \emph{validation} \& \emph{test} costs are estimated on the respective data splits (one could also use other protocols, such as cross-validation with a test set). 
Empirically estimating \emph{validation} and \emph{test} costs induces separate estimation errors.
Validation set quantities are used for \mhpo{} and to approximate the Pareto front. This approximation is then evaluated on the \emph{test} set to obtain an unbiased estimate of the generalization error and also to measure the performance of \mhpo{}.

\begin{figure}[t]
    \centering
    \begin{tabular}{@{\hskip 0mm}c@{\hskip 0mm}c@{\hskip 0mm}c@{\hskip 0mm}}
    (a) & (b) & (c) \\
    \includegraphics[width=.33\textwidth]{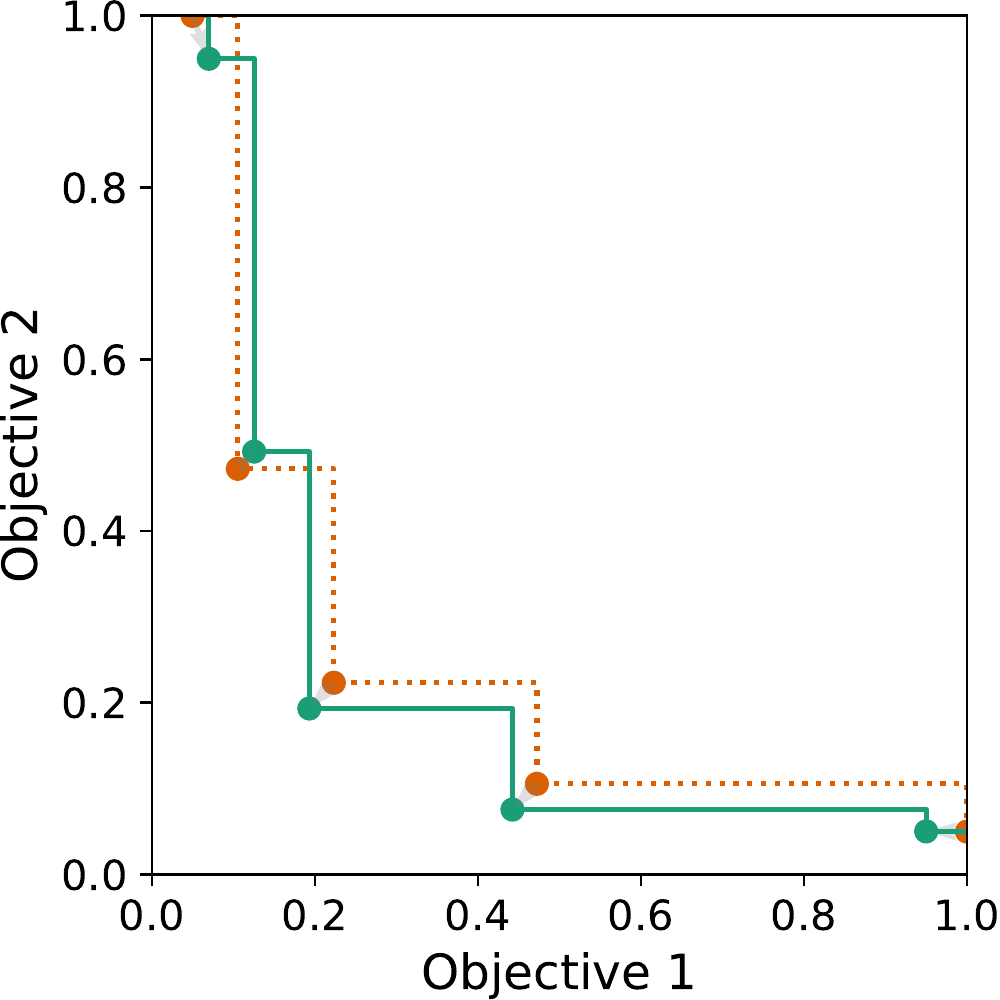} &
    \includegraphics[width=.33\textwidth]{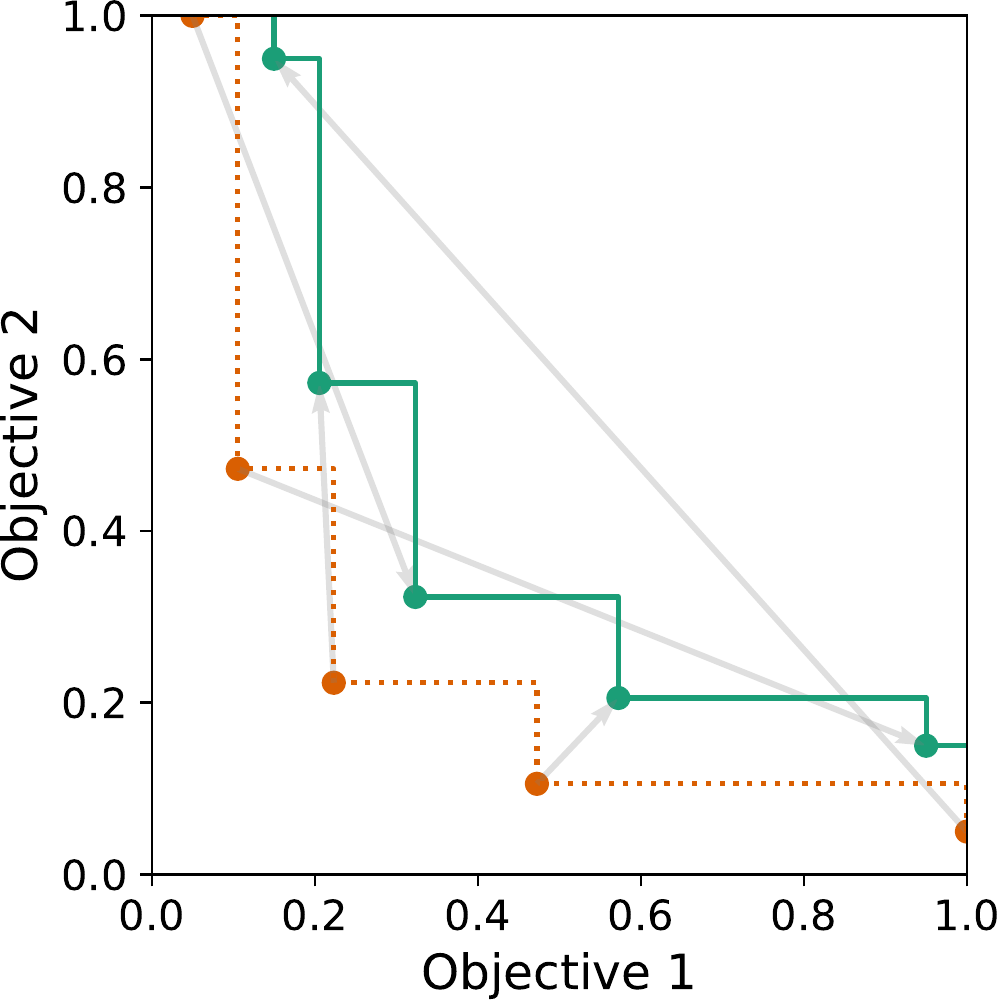} &
    \includegraphics[width=.33\textwidth]{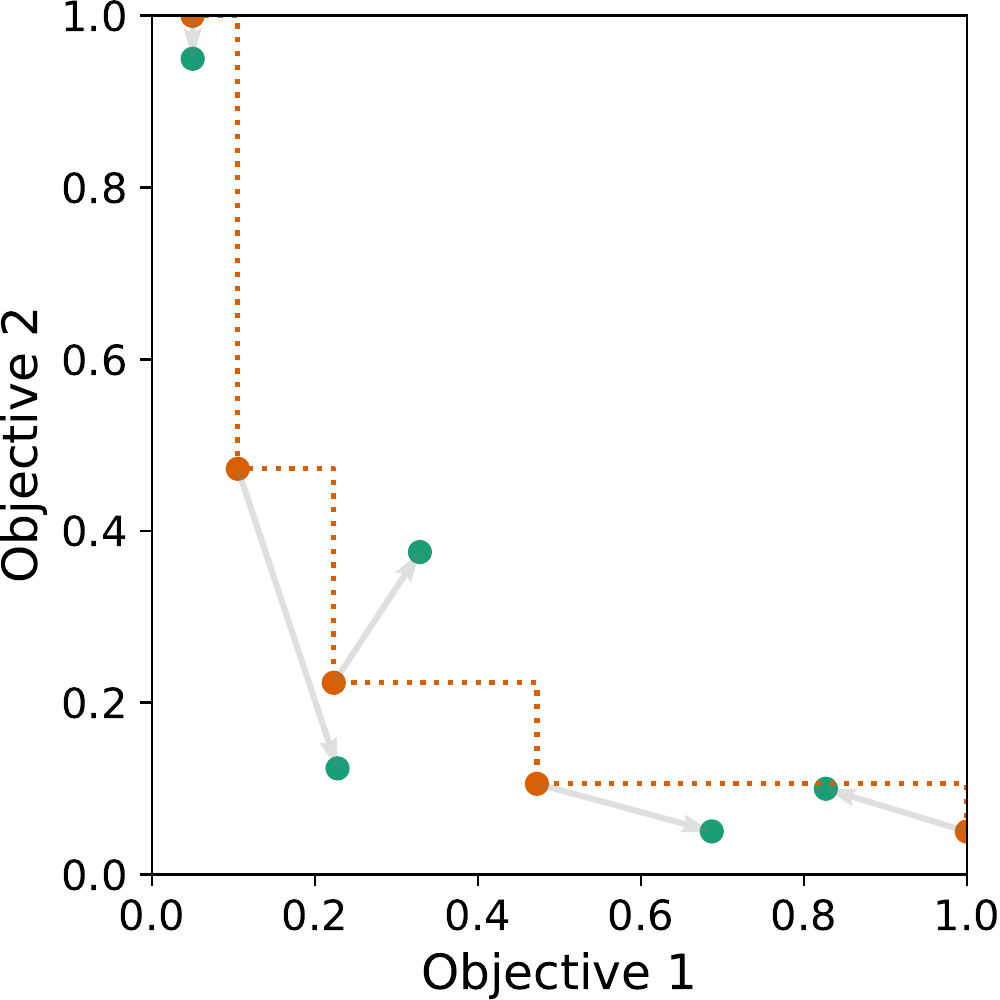} \\
    \multicolumn{3}{c}{\resizebox {0.98\textwidth} {!}{\input{figures/testperf_tikz/legend}}} \\
    \end{tabular}
    \caption{We visualize validation (orange) and test performance (green) of the Pareto set, as found on the validation data set. Considering test performance, (a) all configurations are non-dominated, (b) the configurations are still Pareto-optimal, but switch order, and (c) the configurations are no longer Pareto-optimal.}
    \label{fig:paretofronttestset}
\end{figure}

\textbf{Problems in the Multi-Objective Setting.} Due to these separate estimation errors, an HPC deemed Pareto-optimal on the validation set is not necessarily Pareto-optimal on the test set.\footnote{This is due to a shift in distributions when going from the validation set to the test set due to random sampling. The HPC might then no longer be optimal due to overfitting.} We visualize this in Figure~\ref{fig:paretofronttestset}. On the left-hand side, the Pareto front approximation generalizes well to the test set. 
In the middle, all HPCs are still Pareto-optimal but switch order, which could lead to unexpected performance degradation when selecting an HPC to deploy in practice. However, on the right-hand side, two HPCs are no longer Pareto-optimal, i.e., the Pareto front approximation does not generalize to the test set and contains dominated solutions. 
We would like to highlight that the two problems depicted in the two right-most plots were so far not discussed in the literature, yet, their existence thwarts the evaluation of \mhpo{} algorithms. 

\textbf{A naive solution.} Discarding dominated solutions based on the test set -- which would \emph{not} be possible in practice because we can only access test labels once in the end to compute final performance -- would enable us to compute the hypervolume indicator. For this, we can either consider all evaluated HPCs (as common in assessing the performance of multi-objective methods) or expect the \mhpo{} method to return a reasonable subset (which it believes to be Pareto-optimal). Then, we evaluate these HPCs on the test set, compute the Pareto front approximation based on the test scores, and finally calculate the hypervolume indicator.
Unfortunately, this raises the following two issues. 

\textbf{Issue 1: Overestimation.} We discard dominated points and thus overestimate the true hypervolume of the returned solutions, i.e., ignore solutions that are no longer part of the Pareto set, as displayed in the left-hand-side plot in Figure~\ref{fig:approximationgap}. In practice, a user could pick one of the discarded solutions (based on its validation performance) and observe a worse performance than what we computed as the generalization performance of the optimization method. 

\textbf{Issue 2: Test data leakage.} An adversarial \mhpo{} method could exploit this procedure by returning as many models as possible and thus implicitly selecting its Pareto-optimal set based on the test set, as visualized in the middle of Figure~\ref{fig:paretofronttestset}. While such a system would seemingly obtain a good score, its benefit in practice is limited.

These two issues emphasize the need for a new evaluation protocol that can detect these issues so that we can develop \mhpo{} methods that return reliable Pareto front approximations.

\section{A New Protocol to Measure Generalization}\label{sec:newprotocol}

\begin{figure}[t]
    \begin{tabular}{@{\hskip 0mm}c@{\hskip 0mm}c@{\hskip 0mm}c@{\hskip 0mm}}
    \includegraphics[width=.33\textwidth]{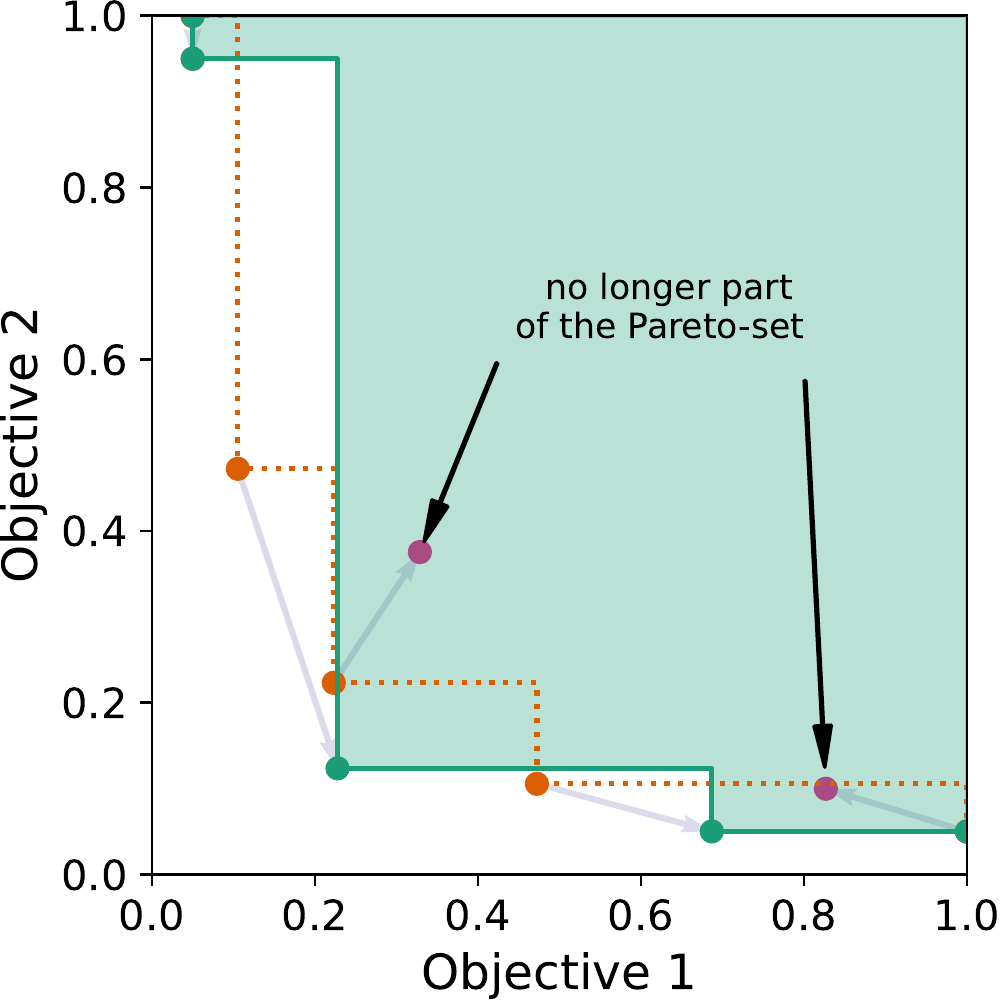} &
    \includegraphics[width=.33\textwidth]{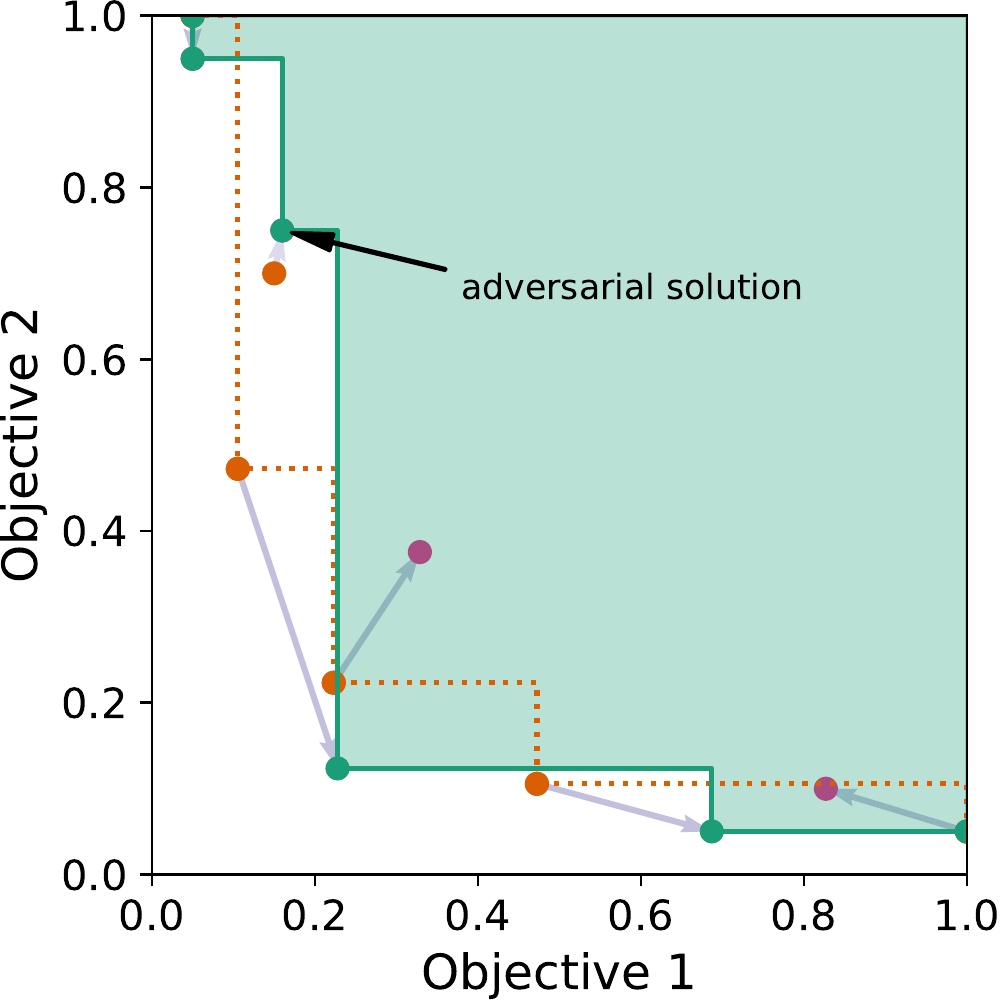} &
    \includegraphics[width=.33\textwidth]{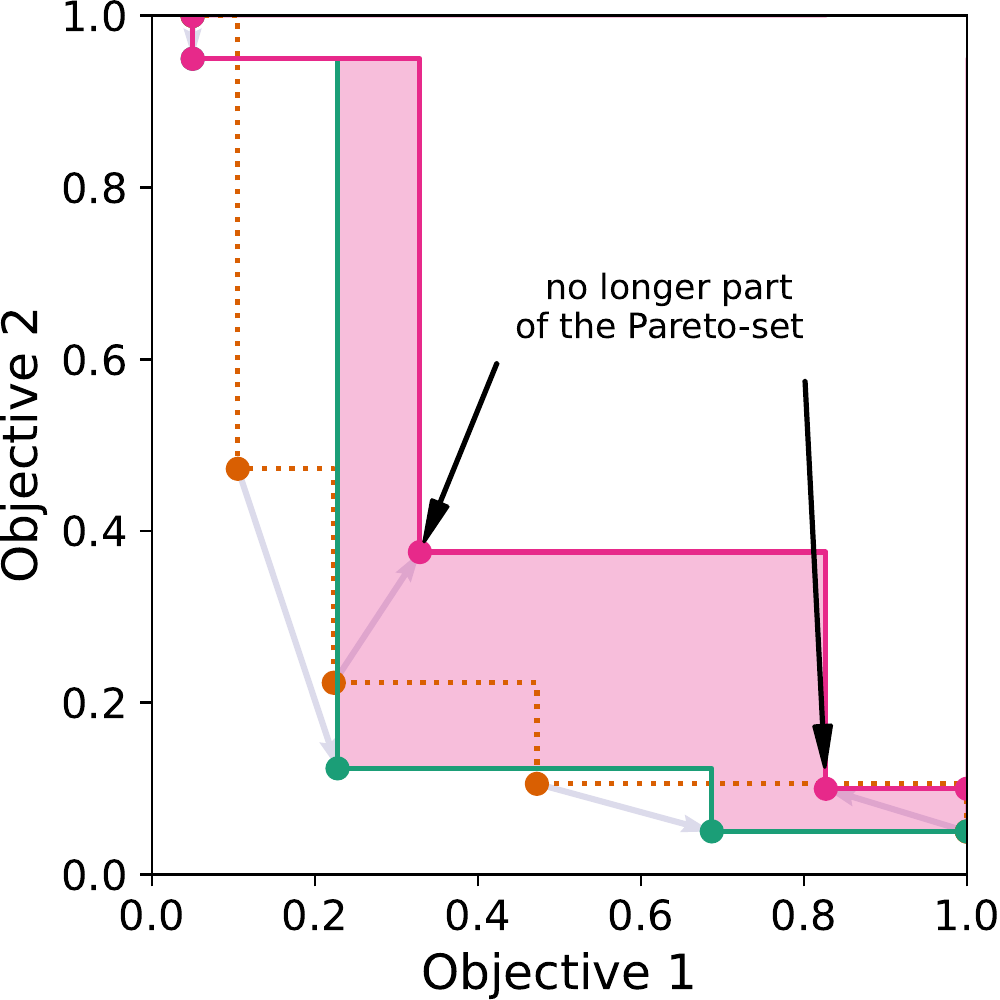} \\
    \multicolumn{3}{c}{\resizebox {0.98\textwidth} {!}{\input{figures/testperf_tikz/legend_gap}}} \\
    \end{tabular}

    \caption{We visualize validation (orange) and test performance (green/pink) of the Pareto set, as found on the validation data set. Left: We show that ignoring dominated points on the test set leads to an overestimation of the hypervolume indicator. Middle: We show how adversarial \mhpo{} can return points that lead to an increased hypervolume on the test data. Right: We show our proposed optimistic Pareto-set (green), pessimistic Pareto-set (pink), and the approximation gap between the optimistic and pessimistic Pareto-set (pink area).}
    \label{fig:approximationgap}
\end{figure}

We propose a new evaluation protocol to assess the performance and robustness of an \mhpo{} method reliably.
We introduce the concept of \textit{optimistic} and \textit{pessimistic} approximations of the Pareto front. We visualize this on the right-hand side of Figure~\ref{fig:approximationgap}. 
Given an approximation of a Pareto front $\tilde{\mathcal{P}}$ that was computed using the validation split of a data set, we formally define the \textit{optimistic} Pareto front as
\begin{equation}
    \mathcal{P}_{optimistic} = \{\val \in \vec{c}(\tilde{\mathcal{P}}) \,|\, \nexists\; \val' \in \vec{c}(\tilde{\mathcal{P}}) \text{ s.t. } \val' \prec_{test} \val \},
\end{equation}
where $\prec_{test}$ denotes a dominance relationship between costs $\val$ and $\val'$ evaluated on the test set instead of the validation set. Similarly, we define the \textit{pessimistic} Pareto front as 
\begin{equation}
    \mathcal{P}_{pessimistic} = \{\val \in \vec{c}(\tilde{\mathcal{P}}) \,|\, \nexists \; \val' \in \vec{c}(\tilde{\mathcal{P}}) \text{ s.t. } \val \prec_{test} \val' \}.
\end{equation}
Then, we can compute the hypervolume for both approximations. The difference between both volumes indicates how robust the Pareto front approximation is when going to test data, and we refer to it as the \emph{approximation gap}. If it is zero, the Pareto set remains identical when moving from validation to test data.
If it is greater than zero, then returned HPCs are dominated on the test data.

We can now compare two \mhpo{} methods, \emph{A} and \emph{B}, based on their hypervolume using the following three criteria: (1)  \emph{hypervolume difference:} by checking if the optimistic estimate of the hypervolume of an \mhpo{} method \emph{A} is smaller than the pessimistic estimate of the hypervolume of an \mhpo{} method \emph{B}, (2)~\emph{dominance:} by using the notion of the optimistic and pessimistic Pareto set to check if pessimistic Pareto front approximation of \emph{A} dominates the optimistic Pareto front approximation of \emph{B}, following the popular idea of Pareto front dominance~\citep{emmerich-naturecomp18a}, and (3) \emph{approximation gap:} by comparing the gap between the optimistic and pessimistic hypervolume across \mhpo{} methods whereas a smaller gap indicates a more robust approximation of the Pareto front.

\section{Experimental Evaluation}\label{sec:experiment}

In this section, we first show that the approximation gap appears in practice and second, experimentally check whether we can now compare algorithms again.

\subsection{Demonstration of Approximation Gap}

\begin{figure}[t]
    \centering
    \includegraphics[width=0.99\textwidth]{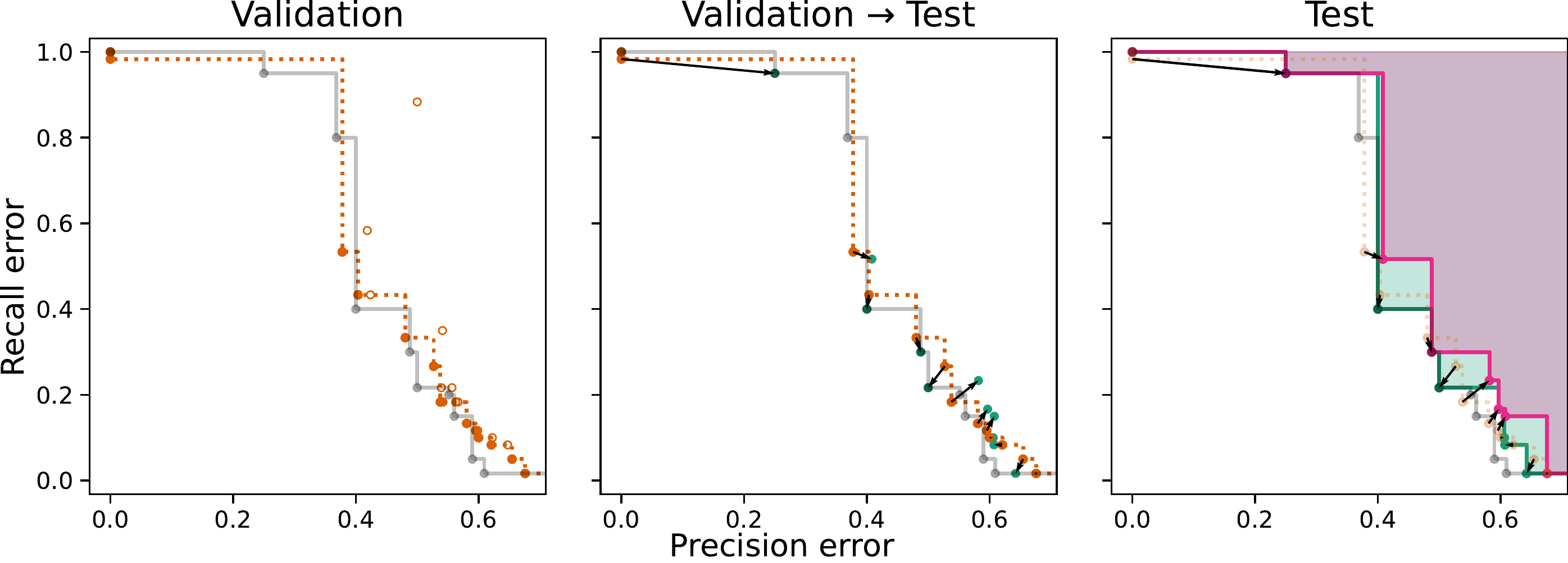}
    {\resizebox {0.99\textwidth} {!}{\input{figures/testperf_tikz/legend_real_data}}}
    \caption{Precision vs Recall. The left plot focuses on the validation error, the middle plot depicts the test error of points from the Pareto set on the validation set, and the right-hand-side plot depicts the approximation of the optimistic and the pessimistic Pareto sets.}
    \label{fig:experiment}
\end{figure}

We first demonstrate the existence of the approximation gap by tuning the hyperparameters of a machine learning algorithm. Concretely, we tune the hyperparameters of a random forest model~\citep{breiman-mlj01a} with $40$ iterations of random search~\citep{bergstra-jmlr12a} on the German credit dataset~\citep{ucimlrepo}. We provide the configuration space and dataset description in Appendix~\ref{app:searchspace}. 
We use precision and recall as objectives, motivated by the fact that both are often ad-hoc combined into the F1 score~\citep{manning-book08a} despite this being an inherently multi-objective problem. Following~\citet{horn-ssci16a}, we tune class weights to account for the unbalanced targets. We split the dataset into 60\% train, 20\% valid, and 20\% test data. For every HPC, we train a single model, record the precision and recall metrics on both the validation and test set and visualize the results in Figure~\ref{fig:experiment}.

The plots are similarly structured as Figures~\ref{fig:paretofronttestset} and~\ref{fig:approximationgap}, and we depict validation performance (in orange) and test performance (in green) \wrt{} both objectives for all evaluated HPCs. The left-hand-side plot highlights the validation losses and the approximation of the Pareto front using the validation set. The middle plot shows how the performance changes when evaluating these HPCs on the test set. Furthermore, we show the hypothetical true Pareto-set $\tilde{\mathcal{P}}_{test}$ on the test data (which we cannot compute in practice; in grey). 
The right-hand-side plot shows the optimistic (in green) and pessimistic Pareto-set (in pink), which we described above. We observe $\tilde{\mathcal{P}}_{test} \prec \mathcal{P}_{optimistic} \prec \mathcal{P}_{pessimistic}$, while perfect generalization to the test set would give us $\tilde{\mathcal{P}}_{test} = \mathcal{P}_{optimistic} = \mathcal{P}_{pessimistic}$.

\subsection{Can We Compare Two Algorithms Again?}

Having seen the approximation gap, we now experimentally check whether the three criteria introduced in Section~\ref{sec:newprotocol} enable comparisons between two different algorithms again. For this, we optimize the hyperparameters of a random forest and a linear classifier with random search for 50, 100, 200, and 500 iterations. We use the same experimental setup and configuration space for the random forest as above and display the configuration space for the linear model in Appendix~\ref{app:searchspace}.

\begin{table}[t]
    \caption{Hypervolume indicators and approximation gap obtained by random search optimizing the hyperparameters of a random forest (top) and a linear model (bottom).}
    \label{tab:comparison}    \centering
    \begin{tabular}{l|l|r|r|r|r}
        \toprule
         & & 50 & 100 & 200 & 500  \\
         \midrule
         \multirow{4}{*}{Random Forest\ } & Validation HV\  & $\ 0.5660$ & $\ 0.6357$ & $\ 0.6426$ & $\ 0.6563$ \\
         & Pessimistic HV\  & $0.5651$ & $0.5972$ & $0.6060$ & $0.5721$ \\
         & Optimistic HV\  & $0.5833$ & $0.6128$ & $0.6272$ & $0.6382$ \\
         & Approximation Gap\  & $0.0181$ & $0.0156$ & $0.0212$ & $0.0661$ \\
         \midrule
         \multirow{4}{*}{Linear Model} & Validation HV\  & $0.5804$ & $0.5943$ & $0.6330$ & $0.6340$ \\
         & Pessimistic HV\  & $0.5970$ & $0.5628$ & $0.5860$ & $0.5641$ \\
         & Optimistic HV\  & $0.5989$ & $0.5798$ & $0.5994$ & $0.5918$ \\
         & Approximation Gap\  & $0.0009$ & $0.0170$ & $0.0134$ & $0.0277$ \\
         \bottomrule
    \end{tabular}
\end{table}

First, we show the hypervolume indicator on the validation set, the pessimistic and optimistic hypervolume indicator, and the approximation gap in Table~\ref{tab:comparison}. As expected, we see that the validation hypervolume increases monotonically with more function evaluations, and after 100 function evaluations, the random forest has a larger validation hypervolume than the linear model, even with 500 function evaluations. Next, we look at the pessimistic and optimistic hypervolume. We can observe that there is no guarantee that they increase together with the validation hypervolume, which means that solutions obtained on the validation set do not generalize to the test set. This can be seen, for example, for the random forest, where the pessimistic hypervolume decreases when going from 200 to 500 function evaluations, while the optimistic hypervolume increases. For the linear model, we can even observe that both the optimistic and pessimistic hypervolume decrease, which can be seen when going from 50 to 100 and from 200 to 500 function evaluations. We can now also compare the two algorithms by comparing the hypervolume indicators (method (1) from Section~\ref{sec:newprotocol}), checking whether the pessimistic hypervolume indicator of one algorithm is larger than the optimistic hypervolume indicator of the other. Using this comparison method, we can conclude (1) that the linear model performs better than the random forest after 50 \mhpo{} function evaluations, (2) that we cannot make a statement about 100 function evaluations, (3) that the random forest is better after 200 function evaluations, and (4) that we cannot make a statement at 500 function evaluations.

\begin{figure}[t]
    \centering
    \includegraphics[width=0.99\textwidth]{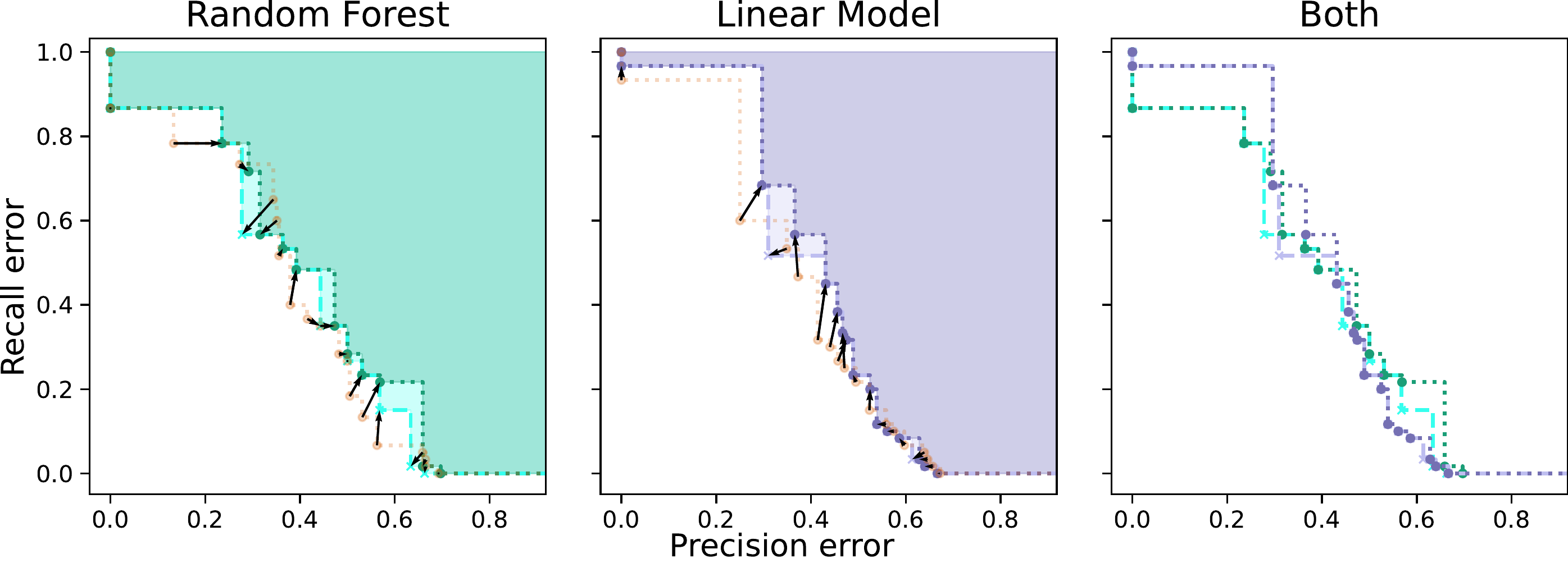}
    {\resizebox {0.99\textwidth} {!}{\input{figures/testperf_tikz/legend_comparison}}}
    \caption{Optimistic and pessimistic Pareto fronts for the random forest (left), the linear model (middle), and both (right) after 200 iterations of random search. For both models, we plot the pessimistic Pareto front in a darker color and using circle markers and the optimistic Pareto front in a lighter color and using star markers; and we use the same colors in the plot on the right-hand side. Furthermore, in the left and middle plots, we also give the validation Pareto front in light orange (similar to Figure~\ref{fig:experiment}).}
    \label{fig:experiment_2}
\end{figure}

Second, we display the Pareto fronts of the two optimized algorithms in Figure~\ref{fig:experiment_2} to check whether one Pareto front dominates the other (method (2) from Section~\ref{sec:newprotocol}). We display results after 200 function evaluations, i.e., when the pessimistic hypervolume of the random forest is higher than the optimistic hypervolume of the linear model. This hypervolume dominance is a necessary but not a sufficient condition for Pareto front dominance. In this case, there are indeed solutions for the linear model (denoted as SGD) that are not dominated by the Pareto front of the random forest, making it impossible to state that one model is generally better than the other.

Finally, we examine the approximation gap of the two \mhpo{} algorithms (method (3) from Section~\ref{sec:newprotocol}). The approximation gap is not a monotonic function. It can decrease when the number of function evaluations of the search algorithm increases (random forest from 50 to 100 function evaluations and linear model from 100 to 200 function evaluations). However, the approximation gap can become quite large as we can observe for the random forest with 500 function evaluations, where its size is 10\% of the optimistic hypervolume. Also, the approximation gap can be larger than any hypervolume measurement over time, and we argue that this makes it impossible to conclude whether the tuned algorithm actually has improved. On a positive note, we also observe that the two different algorithms appear to have different approximation gaps, which suggests that there are machine learning models that provide more stable solutions. 

We conclude this section by answering the question in the section name: yes, the new evaluation protocol allows us to measure the generalization performance of an \mhpo{} algorithm and thereby also to compare two \mhpo{} algorithms or two machine learning models optimized by one \mhpo{} algorithm again.

\section{Prior Evaluation Protocols for Multi-Objective Optimization}\label{sec:prior}

This section reviews how prior works address the problem of measuring generalization performance in a multi-objective setting. We would first like to note that our setup differs from standard optimization problems under noise since we cannot recover the true function value by repeatedly evaluating the function of interest.\footnote{If the true function values of evaluated configurations cannot be recovered due to budget restrictions, our proposed evaluation protocol can be applied as well to deal with solutions that are no longer part of the Pareto front on the test set.} However, in our case, the performance of a model selected on the validation set suffers from a distribution shift on the test set. We are unaware of a method for describing such distribution shifts that happen when moving from the validation to the test set.\footnote{Distributionally Robust Bayesian Optimization~\citep{kirschner-aistats20a} is an algorithm that could be used in such a setting and the paper introducing it explicitly states \automl{} as an application, but does neither demonstrate its applicability to \automl{} nor elaborates on how to describe the distribution shift in a way the algorithm could handle it.}

To the best of our knowledge, no one has yet explicitly studied how to measure the generalization error of HPCs for machine learning models in a multi-objective setting. We found two works that evaluate multi-objective generalization: \citet{horn-emo17a} use the naive protocol we outline above, and \citet{binder-gecco20a} solely compute, what we call, the optimistic Pareto front. Nonetheless, we would like to emphasize that these works employ these measures in an ad-hoc fashion without any discussion or justification. In contrast, we thoroughly introduce the approximation gap and the concepts behind it.
In the field of \mhpo{}, we found that researchers so far use scalarization to choose a final model to evaluate~\citep{cruz-icdm21a}, pick a model based on a single metric~\citep{gardner-dsaa19a}, or use handcrafted heuristics to select a final model~\citep{feffer-arxiv22a}.

A similar problem exists for constrained optimization: a solution that satisfies the constraints on the validation set can violate the constraint on the test set. \citet{lobato-jmlr16a} found that "When the constraints are noisy, reporting the best observation is an overly optimistic metric because the best feasible observation might be infeasible in practice" and evaluate a "ground-truth score" by evaluating the final recommendation multiple times, treating a constrained violation as $100\%$ classification error. They tuned a 
neural network on MNIST under inference time constraints and tuned Hamiltonian Monte Carlo under the constraint that the generated samples pass convergence diagnostic tests.
In the field of noisy constrained Bayesian optimization, researchers have suggested an \emph{identification step} to select the best point after optimization~\citep{gelbart-uai14a}, and~\citet{letham-ba18a} study the proportion of replicates in which the proposed method manages to find suitable solutions, but without scalarizing the final objective as done by~\cite{lobato-jmlr16a}.

Moreover, for the problem of multi-objective ranking and selection (identification of the Pareto set from a finite set of choices), the F1 metric was proposed for judging the final result~\citep{gonzalez-arxiv22a}. However, this does not quantify the solution quality in performance space. 
Last, the terminology of optimistic and pessimistic Pareto set has also been used in the context of approximating a Pareto front from the predictions of a probabilistic model~\citep{iqbal-arxiv20a}.

\section{Conclusions and Future Work}\label{sec:conclusion}

We have demonstrated that the standard evaluation protocol for single-objective HPO is inapplicable in the multi-objective setting and, as a remedy, introduced optimistic and pessimistic Pareto sets. Based on these, we can compare multi-objective algorithms using the \emph{hypervolume difference}, \emph{dominance}, or the new \emph{approximation gap}.
Furthermore, we can detect if the MHPO algorithm leads to an unstable solution, i.e., a large approximation gap, the analogue to over-fitting in single-objective optimization. In an experimental study, we have verified the existence of the approximation gap and demonstrated that we can now compare two machine learning models optimized for multiple metrics again.

In the future, we plan to 
(1) measure the effect of this problem over a large number of datasets and varying numbers of function evaluations, (2) extend our analysis to take measurement noise into account and (3) extend our protocol to multiple repetitions and cross-validation. Furthermore, we want to (4) evaluate additional multi-objective problems, e.g., trading off true-positive rates and false-positive rates~\citep{levesque-gecco12a,horn-ssci16a,karl-arxiv22a} or fairness and predictive performance~\citep{chakraborty-ase19a,cruz-icdm21a,schmucker-arxiv21a} and (5) study the related problem of distribution shifts in data streams.

\subsection*{Acknowledgements}
Robert Bosch GmbH is acknowledged for financial support. Also, this research was partially supported by TAILOR, a project funded by EU Horizon 2020 research and innovation programme under GA No 952215.
The authors of this work take full responsibility for its content.

\bibliography{strings,lib,local,proc}
\bibliographystyle{splncs04nat}


\appendix

\section{Experimental Details}\label{app:searchspace}

\vspace{-10pt}

\begin{table}[h]
    \centering
    \label{tab:searchspace}
    \begin{tabular}{lc|lc}
        \toprule
        \multicolumn{2}{c}{Random Forest} & \multicolumn{2}{c}{Linear Model} \\
        Hyperparameter name & Search space & Hyperparameter name & Search Space \\
        \midrule
        criterion & [gini, entropy] & penalty & [l2, l1, elasticnet] \\
        bootstrap & [True, False] & alpha & $[1e-6, 1e-2]$, log \\
        max\_features & $[0.0, 1.0]$ & l1 ratio & $[0.0, 1.0]$ \\
        min\_samples\_split & $[2, 20]$ & fit\_intercept & [True, False] \\
        min\_samples\_leaf & $[1, 20]$ & eta0 & $[1e-7, 1e-1]$ \\
        pos\_class\_weight exponent & $[-7, 7]$ & pos\_class\_weight exp. & $[-7, 7]$ \\
        \bottomrule
    \end{tabular}
\end{table}

We provide the random forest and linear model search spaces in Table~\ref{tab:searchspace}. We fit the linear model with stochastic gradient descent and use an \emph{adaptive} learning rate and minimize the log loss (please see the scikit-learn~\citep{scikit-learn} documentation for a description of these). Because we are dealing with unbalanced data, we consider the class weights as a hyperparameter and tune the weight of the minority (positive) class in the range of $[2^{-7}, 2^7]$ on a log-scale~\citep{konen-acm11a,horn-ssci16a}.
To deal with categorical features, we use one hot encoding. We transform the features for the linear models using a quantile transformer with a normal output distribution.

We use the German credit dataset~\citep{ucimlrepo} because it is relatively small, leading to high variance in the algorithm performance, and unbalanced. We downloaded the dataset from OpenML~\citep{vanschoren-sigkdd14a} using the OpenML-Python API~\citep{feurer-jmlr21a} as task ID $31$, but conducted our own 60/20/20 split. 
It is a binary classification problem with 30\% positive samples. The dataset has 1000 samples and 20 features. Out of the 20 features, 13 are categorical. The dataset contains no missing values.

\end{document}

%% file: macros.tex
\newcommand{\ournote}[1]{
	\noindent~\\
	\fcolorbox{red}{orange}{\parbox{0.99\columnwidth}{#1}}
}

\newcommand{\bb}{black-box}
\newcommand{\mf}{multi-fidelity}
\newcommand{\mhpo}{MHPO}

\newcommand{\paretosetsize}{p}
\newcommand{\metric}{m}
\newcommand{\pipeline}{\mathcal{M}}
\newcommand{\costdim}{M}
\newcommand{\metricdim}{\costdim}

\newcommand{\dataset}{\mathcal{D}}
\newcommand{\inp}{\vec{x}}
\newcommand{\out}{y}

\newcommand{\wrt}{w.r.t.}

\newcommand{\configspace}{configuration space}
\newcommand{\bestfound}{\conf^+}
\newcommand{\dimpcs}[0]{d}
\newcommand{\pcs}[0]{\bm{\Lambda}}
\newcommand{\pcsdim}[0]{n}
\newcommand{\seeds}[0]{\vec{s}}
\newcommand{\seed}[0]{s}
\newcommand{\cost}[0]{c}
\newcommand{\costs}[0]{\vec{c}}
\newcommand{\normal}[2]{\mathcal{N}(#1, #2)}
\newcommand{\lognormal}[2]{log\mathcal{N}(#1, #2)}
\newcommand{\expected}{\mathbb{E}}
\newcommand{\ind}{I}
\newcommand{\real}{\mathbb{R}}

\newcommand{\insts}[0]{\Pi}
\newcommand{\instD}[0]{\mathcal{D}_\insts}
\newcommand{\conf}[0]{\bm{\lambda}}
\newcommand{\configspacelambda}[0]{\bm{\Lambda}}
\newcommand{\cutoff}[0]{\kappa}
\newcommand{\loss}[0]{\mathcal{L}}

\renewcommand{\vec}[1]{\bm{#1}}
\newcommand{\algo}[0]{\mathcal{A}}
\newcommand{\portfolio}[0]{\mathcal{P}}
\newcommand{\feats}[0]{\bm{f}}
\newcommand{\allfeats}[0]{\mathcal{F}}
\newcommand{\predictor}[0]{\hat{m}}

\newcommand{\bo}{\textit{BO}}
\newcommand{\smac}{\textit{SMAC}}
\newcommand{\random}{\textit{RS}}

\newcommand{\scikit}{\textit{scikit-learn}}
\newcommand{\pytorch}{\textit{PyTorch}}
\newcommand{\automl}{AutoML}
\newcommand{\fautoml}{fairness-aware \automl{}}
\newcommand{\Fautoml}{Fairness-aware \automl{}}

\newcommand{\NNewDatasets}{14}

\newcommand{\val}[0]{\bm{\zeta}}

%% file: figures/testperf_tikz/legend.tex
\begin{tikzpicture}

\definecolor{chocolate217952}{RGB}{217,95,2}
\definecolor{darkcyan27158119}{RGB}{27,158,119}
\definecolor{darkgray176}{RGB}{176,176,176}
\definecolor{deeppink23141138}{RGB}{231,41,138}
\definecolor{lightgray204}{RGB}{204,204,204}
\definecolor{lightslategray117112179}{RGB}{117,112,179}

\begin{axis}[
hide axis,
legend cell align={left},
legend style={fill opacity=0.8, draw opacity=1, text opacity=1,
legend columns=4,legend style={draw=none,column sep=5ex},
legend cell align={left},
draw=lightgray204},
tick align=outside,
tick pos=left,
x grid style={darkgray176},
xlabel={unfairness},
xmin=0, xmax=1,
xtick style={color=black},
y grid style={darkgray176},
ylabel={error},
ymin=0, ymax=1,
ytick style={color=black}
]


\addlegendimage{semithick, chocolate217952, const plot mark right, dotted, mark=*, mark size=3, mark options={solid}}
\addlegendentry{\large Pareto-set (validation)}

\addlegendimage{semithick, darkcyan27158119, only marks, mark=*, mark size=3, mark options={solid}}
\addlegendentry{\large Performance (test)}




\addlegendimage{thick, black!50, ->, mark size=5, mark options={solid}}
\addlegendentry{\large Shift val. $\rightarrow$ test}

\end{axis}

\end{tikzpicture}

%% file: figures/testperf_tikz/legend_gap.tex
\begin{tikzpicture}

\definecolor{chocolate217952}{RGB}{217,95,2}
\definecolor{darkcyan27158119}{RGB}{27,158,119}
\definecolor{darkgray176}{RGB}{176,176,176}
\definecolor{deeppink23141138}{RGB}{231,41,138}
\definecolor{lightgray204}{RGB}{204,204,204}
\definecolor{lightslategray117112179}{RGB}{117,112,179}

\begin{axis}[
hide axis,
legend cell align={left},
legend style={fill opacity=0.8, draw opacity=1, text opacity=1,
legend columns=3,legend style={draw=none,column sep=5ex,cells={align=left}},
legend cell align={left},
draw=lightgray204},
tick align=outside,
tick pos=left,
x grid style={darkgray176},
xlabel={unfairness},
xmin=0, xmax=1,
xtick style={color=black},
y grid style={darkgray176},
ylabel={error},
ymin=0, ymax=1,
ytick style={color=black}
]


\addlegendimage{semithick, chocolate217952, const plot mark right, dotted, mark=*, mark size=3, mark options={solid}}
\addlegendentry{\large Pareto-set (validation)}

\addlegendimage{semithick, darkcyan27158119, only marks, mark=*, mark size=3, mark options={solid}}
\addlegendentry{\large Performance (test)}

\addlegendimage{semithick, darkcyan27158119, const plot mark right, mark=*, mark size=3, mark options={solid}}
\addlegendentry{\large (Optimistic) Pareto-set}

\addlegendimage{thick, black!50, ->, mark size=5, mark options={solid}}
\addlegendentry{\large Shift val. $\rightarrow$ test}

\addlegendimage{area legend, draw=darkcyan27158119, fill=darkcyan27158119, opacity=0.3}
\addlegendentry{\large Overestimated Hypervolume}

\addlegendimage{semithick, deeppink23141138, const plot mark right, mark=*, mark size=3, mark options={solid}}
\addlegendentry{\large Pessimistic Pareto-set}

\addlegendimage{area legend, draw=deeppink23141138, fill=deeppink23141138, opacity=0.3}
\addlegendentry{\large Diff optimistic/  pessimistic Pareto-set}

\end{axis}

\end{tikzpicture}

%% file: figures/testperf_tikz/legend_real_data.tex
\begin{tikzpicture}

\definecolor{chocolate217952}{RGB}{217,95,2}
\definecolor{darkcyan27158119}{RGB}{27,158,119}
\definecolor{darkgray176}{RGB}{176,176,176}
\definecolor{deeppink23141138}{RGB}{231,41,138}
\definecolor{lightgray204}{RGB}{204,204,204}
\definecolor{lightslategray117112179}{RGB}{117,112,179}
\definecolor{green}{RGB}{102,166,30}

\begin{axis}[
hide axis,
legend cell align={left},
legend style={fill opacity=0.8, draw opacity=1, text opacity=1,
legend columns=4,legend style={draw=none,column sep=5ex},
legend cell align={left},
draw=lightgray204},
tick align=outside,
tick pos=left,
x grid style={darkgray176},
xlabel={unfairness},
xmin=0, xmax=1,
xtick style={color=black},
y grid style={darkgray176},
ylabel={error},
ymin=0, ymax=1,
ytick style={color=black},
num1/.style={chocolate217952,only marks,mark=o},
num2/.style={darkcyan27158119,only marks,mark=*},
combo legend/.style={
          legend image code/.code={
            \draw [/pgfplots/num1] plot coordinates {(1mm,0cm)};
            \draw [/pgfplots/num2] plot coordinates {(4.5mm,0cm)};
          }
        }
]

\addlegendimage{combo legend}
\addlegendentry{\large Performance (val/test)}



\addlegendimage{semithick, chocolate217952, const plot mark right, dotted, mark=*, mark size=3, mark options={solid}}
\addlegendentry{\large Pareto-set (val)}

\addlegendimage{semithick, black, const plot mark right, mark=*, mark size=3, mark options={solid}, opacity=0.3}
\addlegendentry{\large Pareto-set (test)}

\addlegendimage{area legend, draw=deeppink23141138, fill=deeppink23141138, opacity=0.3}
\addlegendentry{\large Pessimistic Hypervolume}

\addlegendimage{area legend, draw=darkcyan27158119, fill=darkcyan27158119, opacity=0.3}
\addlegendentry{\large Optimistic Hypervolume}

\addlegendimage{semithick, darkcyan27158119, const plot mark right, mark=*, mark size=3, mark options={solid}}
\addlegendentry{\large Optimistic Pareto-set}

\addlegendimage{semithick, deeppink23141138, const plot mark right, mark=*, mark size=3, mark options={solid}}
\addlegendentry{\large Pessimistic Pareto-set}

\addlegendimage{thick, black!50, ->, mark size=5, mark options={solid}}
\addlegendentry{\large Shift val. $\rightarrow$ test}

\end{axis}

\end{tikzpicture}

%% file: figures/testperf_tikz/legend_comparison.tex
\begin{tikzpicture}

\definecolor{chocolate217952}{RGB}{217,95,2}
\definecolor{darkcyan27158119}{RGB}{27,158,119}
\definecolor{darkgray176}{RGB}{176,176,176}
\definecolor{darkergray50}{RGB}{50,50,50}
\definecolor{deeppink23141138}{RGB}{231,41,138}
\definecolor{lightcyan54255238}{RGB}{54,255,238}
\definecolor{lightgray204}{RGB}{204,204,204}
\definecolor{lightslategray117112179}{RGB}{117,112,179}
\definecolor{green}{RGB}{102,166,30}
\definecolor{purple117112179}{RGB}{117,112,179}
\definecolor{lightpurple190190240}{RGB}{190,190,240}

\pgfdeclareplotmark{mystarcyan}{
    \node[star, star points=5, star point ratio=0.5, draw=lightcyan54255238, solid, fill=lightcyan54255238, minimum width=3pt, inner sep=0pt, outer sep=0pt, anchor=center] {};
}
\pgfdeclareplotmark{mystarpurple}{
    \node[star, star points=5, star point ratio=0.5, draw=lightpurple190190240, solid, fill=lightpurple190190240, minimum width=3pt, inner sep=0pt, outer sep=0pt, anchor=center] {};
}

\begin{axis}[
hide axis,
legend cell align={left},
legend style={fill opacity=0.8, draw opacity=1, text opacity=1,
legend columns=3,legend style={draw=none,column sep=5ex},
legend cell align={left},
draw=lightgray204},
tick align=outside,
tick pos=left,
x grid style={darkgray176},
xlabel={unfairness},
xmin=0, xmax=1,
xtick style={color=black},
y grid style={darkgray176},
ylabel={error},
ymin=0, ymax=1,
ytick style={color=black},
num1/.style={darkcyan27158119,only marks,mark=*},
num2/.style={lightcyan54255238,only marks,mark=mystarcyan},
num3/.style={purple117112179,only marks,mark=*},
num4/.style={lightpurple190190240,only marks,mark=mystarpurple},
combo legend1/.style={
          legend image code/.code={
            \draw [/pgfplots/num1] plot coordinates {(1mm,0cm)};
            \draw [/pgfplots/num2] plot coordinates {(4.5mm,0cm)};
          }
        },
combo legend2/.style={
          legend image code/.code={
            \draw [/pgfplots/num3] plot coordinates {(1mm,0cm)};
            \draw [/pgfplots/num4] plot coordinates {(4.5mm,0cm)};
          }
        }
]

\addlegendimage{semithick, chocolate217952, const plot mark right, dotted, mark=*, mark size=3, mark options={solid}, opacity=0.25}
\addlegendentry{\large Pareto-set (val)}

\addlegendimage{combo legend1}
\addlegendentry{\large Pareto-set Random Forest (val/test)}

\addlegendimage{combo legend2}
\addlegendentry{\large Pareto-set Linear Model (val/test)}

\addlegendimage{thick, black!50, ->, mark size=5, mark options={solid}}
\addlegendentry{\large Shift val. $\rightarrow$ test}

\addlegendimage{area legend, draw=darkcyan27158119, fill=darkcyan27158119, opacity=1.0}
\addlegendentry{\large Pessimistic Hypervolume Random Forest}

\addlegendimage{area legend, draw=lightcyan54255238, fill=lightcyan54255238, opacity=1.0}
\addlegendentry{\large Optimistic Hypervolume Random Forest}

\addlegendimage{area legend, draw=purple117112179, fill=purple117112179, opacity=1.0}
\addlegendentry{\large Pessimistic Hypervolume Linear Model}

\addlegendimage{area legend, draw=lightpurple190190240, fill=lightpurple190190240, opacity=1.0}
\addlegendentry{\large Optimistic Hypervolume Linear Model}

\end{axis}

\end{tikzpicture}